\documentclass[conference]{IEEEtran}
\IEEEoverridecommandlockouts

\usepackage{cite}
\usepackage{amsmath,amssymb,amsfonts}
\usepackage{algorithmic}
\usepackage{graphicx}
\usepackage{textcomp}
\usepackage{xcolor}
\usepackage{multirow}
\usepackage{slashbox}
\usepackage{url,hyperref,lineno,micro type,subcaption}

\usepackage{booktabs}
\usepackage{floatrow}
\usepackage{caption}
\usepackage{wasysym}
\usepackage{wrapfig}

\def\BibTeX{{\rm B\kern-.05em{\sc i\kern-.025em b}\kern-.08em
    T\kern-.1667em\lower.7ex\hbox{E}\kern-.125emX}}
\begin{document}

\title{Brain-Inspired Spiking Neural Network for Online Unsupervised Time Series Prediction

}
\author{\IEEEauthorblockN{Biswadeep Chakraborty}
\IEEEauthorblockA{
\textit{Georgia Institute of Technology}\\
Atlanta, USA \\
biswadeep@gatech.edu}
\and
\IEEEauthorblockN{Saibal Mukhopadhyay}
\IEEEauthorblockA{
\textit{Georgia Institute of Technology}\\
Atlanta, USA \\
saibal.mukhopadhyay@ece.gatech.edu}
}

\maketitle

\begin{abstract}
 
Energy and data-efficient online time series prediction for predicting evolving dynamical systems are critical in several fields, especially edge AI applications that need to update continuously based on streaming data. However, current Deep Neural Network (DNN)-based supervised online learning models require a large amount of training data and cannot quickly adapt when the underlying system changes. Moreover, these models require continuous retraining with incoming data making them highly inefficient. We present a novel Continuous Learning-based Unsupervised Recurrent Spiking Neural Network Model (CLURSNN), trained with spike timing dependent plasticity (STDP) to solve these issues. CLURSNN makes online predictions by reconstructing the underlying dynamical system using Random Delay Embedding by measuring the membrane potential of neurons in the recurrent layer of the recurrent spiking neural network (RSNN) with the highest betweenness centrality. We also use topological data analysis to propose a novel methodology using the Wasserstein Distance between the persistent homologies of the predicted and observed time series as a loss function. We show that the proposed online time series prediction methodology outperforms state-of-the-art DNN models when predicting an evolving Lorenz63 dynamical system.

\end{abstract}

\begin{IEEEkeywords}
spiking neural network, recurrent, STDP, Wasserstein distance, persistent homologies, online time series prediction
\end{IEEEkeywords}

\section{Introduction}
\label{sec:intro}
\begin{figure*}
    \centering
    \includegraphics[width = \textwidth]{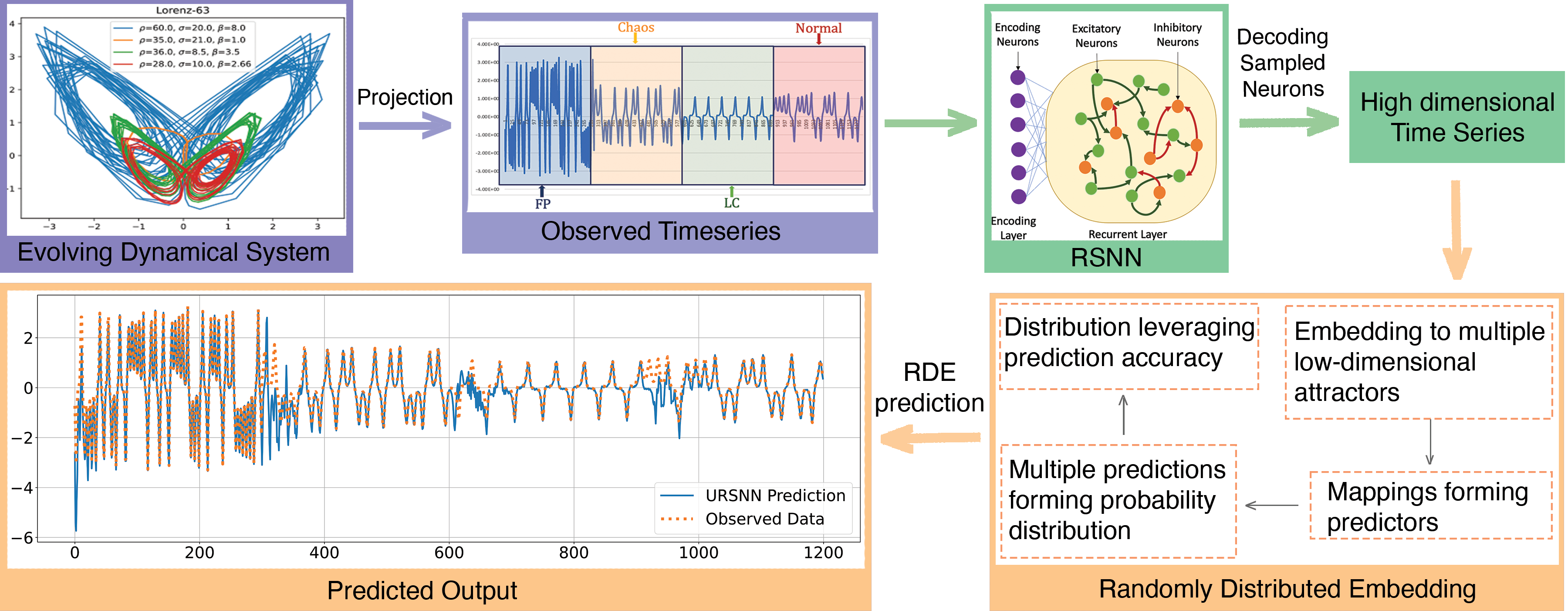}
    \caption{Block Diagram showing the Different Steps for Unsupervised Prediction of Time Series using RSNN}
    \label{fig:diagram}
\end{figure*}

Real-world systems are not static and keep evolving with time, thus creating the need for an evolving learning system \cite{leite2020overview}. Therefore, developing online learning methods that can make predictions by extracting the underlying dynamics from the observed time series for real-time learning and prediction of time-varying environments for machine learning (ML) models running at the edge is critical \cite{chang2021survey}, \cite{cui2018survey}. However, current supervised learning models predict future timesteps based on past observations without any knowledge of the dynamics of the underlying dynamical system  \cite{petropoulos2022forecasting}, \cite{bhatnagar2021merlion}, \cite{li2019enhancing}. Hence, these models fail dramatically for systems continuously evolving, especially systems with sudden data distribution changes. This is because these models require a large amount of data for training and cannot adapt quickly to emergent internal dynamics.

In this paper, we propose a completely unsupervised prediction of the time series, which is quick and robust in adapting to new unseen dynamics. We use recurrent spiking neural networks for extremely-energy efficient brain-inspired models that can continually learn from streaming incoming data using unsupervised plasticity rules. We focus on a recurrent spiking neural network (RSNN) model trained using unsupervised spike timing dependent plasticity rules (STDP) \cite{pool2011spike}, \cite{chakraborty2021characterization} to capture the complex temporal correlations. Recent works \cite{pool2011spike}, \cite{chakraborty2022heterogeneous}, \cite{chakraborty2023heterogeneous} have shown such brain-inspired recurrent spiking neural networks trained with STDP to exhibit promising performance with very few computations. Hence the model continually learns representations of the underlying dynamical systems from which the data is generated. We predict by reconstructing the underlying dynamical systems using random delay embedding on the neuron activation time series data. This method of unsupervised forecasting of time series data contrasts with current regression-type methods used to predict time series data. It is not only data efficient but also helps in the continual modeling of evolving intelligent agents. 
The overall model can be described using the following four-step process:
\begin{enumerate}
    \item We observe the projected time series from an evolving dynamical system. The time series is then fed into the encoder layer of an RSNN, where the data is converted to spike streams that are, in turn, fed into the RSNN, a recurrently connected network of Leaky Integrate and Fire (LIF) neurons, with the synapses being continually updated using STDP. 
    \item We sample neurons from the RSNN with the highest betweenness centrality $\mathcal{C}_b$, and each of the observed spike streams is decoded to get a high-dimensional time series data
    \item We then use the randomly distributed embedding \cite{ma2018randomly} method to reconstruct the underlying dynamical system and predict the system's future steps from this reconstruction.
\end{enumerate}
The detailed process is shown in Fig. \ref{fig:diagram} and explained in greater detail in Section \ref{sec:general}.
The key contributions of this work are as follows:
\begin{itemize}
    \item We propose a completely unsupervised online learning model using a spiking neural network using continual reconstruction of the underlying dynamical system
    \item We sample the nodes with the highest $\mathcal{C}_b$ for the reconstruction and show that such sampling is better than using all the neurons. We relate this sampling to the bottleneck layer of the autoencoder.
    \item We proposed two different methodologies for the reconstruction - one where the model is reconstructed using the RMSE error as the metric and the other where the model is reconstructed using the 1-Wasserstein distance $d_{W,1}$ between the persistent homologies between the predicted ($\hat{p}_H$) and observed time series ($p_H$) as the metric. We show that the Wasserstein-based reconstruction method always outperforms the root mean square error(RMSE)-based reconstruction method, highlighting that the persistent homologies can better capture the underlying dynamics of the system.
    \item We showed that for a limited amount of data, the RSNN-based models outperform the DNN-based models and converge faster. However, with multiple repetitions of the same data, the DNNs outperform the RSNN models.
\end{itemize}

We evaluated the models on a synthetic evolving dynamical system dataset and real-world datasets like YReal, and Dow Jones Industrial Average. The proposed CLURSNN methods outperform the supervised DNN models for online time series prediction tasks. The rest of the paper is organized as follows: Sections \ref{sec:II} and \ref{sec:IV}  discusses the Background and the Methods used in the paper, while in Section \ref{sec:V}, we discuss the experiments performed and the results observed. Finally, we present the conclusions from these observations in Section \ref{sec:VI}.

\section{Background} \label{sec:II}

\textbf{Related Works: }  Recent works on predictive models \cite{petropoulos2022forecasting}, \cite{bhatnagar2021merlion}, \cite{li2019enhancing}, \cite{yue2022ts2vec} demonstrate the ability to discover hierarchical latent representations and complex dependencies. However, such studies focus on batch learning settings where the entire training data set should be available a priori, meaning that the relationship between inputs and outputs is always static. This assumption is restrictive for real-world applications where data arrives in streams, and the underlying dynamical system generating the data changes \cite{gama2014survey}.   Cui et al. \cite{cui2018survey} described the potential of ML methods for various applications like traffic profiling, device identification, system security, Internet of Things applications, edge computing, etc. However, training an artificial intelligence model at the edge from scratch can be time-consuming and energy-inefficient. Therefore, it is desirable to train deep predictors online using only new samples to capture the changing dynamics of the environment \cite{anava2013online}, \cite{liu2016online}.

Despite the ubiquity of online learning in many real-world applications, training detailed predictors online remains a challenge. First, standard deep neural networks converge slowly on the data stream due to the unavailability of offline training advantages such as mini-batch and multi-epoch training \cite{sahoo2017online}, \cite{aljundi2019task}. Overall, deep neural networks have powerful representation learning capabilities but lack mechanisms for successfully training data streams. Therefore, online time series forecasting using deep models is a promising but challenging problem. Several methods have been proposed to solve the online learning problem. However, current deep learning-based models require a huge amount of data for training and are also extremely energy inefficient, making it difficult to deploy in edge devices. 
Brain-inspired spiking neural networks have been proposed as the next generation of energy-efficient neural network models that process information in the discrete spike domain, unlike standard DNNs\cite{kim2022moneta}. Most recent works on spiking neural networks are targeted for time series or spatiotemporal classification \cite{she2021sequence}, \cite{chakraborty2022heterogeneous}, \cite{chakraborty2021fully}. Some recent works also use spiking neural networks for time series forecasting \cite{reid2014financial}, \cite{chakraborty2023heterogeneous}. However, these models are trained using supervised learning methods \cite{reid2014financial} or need a supervised readout layer \cite{margin2022deep}, hence cannot adapt to evolving systems. However, no work exists that uses a completely unsupervised time series prediction method that reconstructs the underlying dynamical system, which is essential for edge ML applications.

\begin{figure}
    \centering
    \includegraphics[width=\columnwidth]{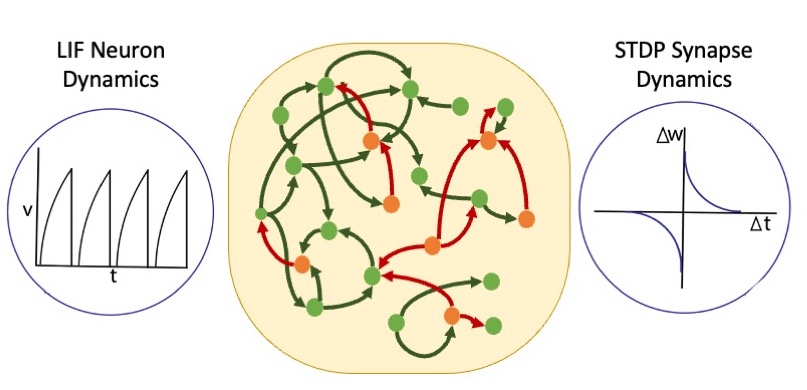}
    \caption{Figure showing the Recurrent Spiking Neural Network with LIF neurons and STDP synapses. The inset figures show the dynamics of the LIF neurons and the STDP weight dynamics.}
    \label{fig:rsnn}
\end{figure}

\textbf{Recurrent Spiking Neural Network (RSNN) } consists of spiking neurons connected with synapses. Our simulations use the Leaky Integrate and Fire (LIF) neuron model. In the LIF neuron model, the membrane potential of the $i$-th neuron $u_{i}(t)$ varies over time as:
\begin{equation}
\tau_{m} \frac{d v_{i}(t)}{d t} =-\left(v_{i}(t)-v_{rest}\right)+I_{i}(t)
\label{eq:1}
\end{equation}
where $\tau_{\mathrm{m}}$ is the membrane time constant, $v_{rest}$ is the resting potential and $I_{\mathrm{i}}$ is the input current. When the membrane potential reaches the threshold value $v_{\text {th }}$ a spike is emitted, $v_{i}(t)$ resets to the reset potential $v_{\mathrm{r}}$ and then enters a refractory period where the neuron cannot spike. Spikes emitted by the $j$th neuron at a finite set of times $\left\{t_{j}\right\}$ can be formalized as a spike train $\displaystyle S_{i}(t)=\sum \delta\left(t-t_{i}\right)$. We use a STDP rule \cite{pool2011spike} for updating a synaptic weight ($\Delta w$) and is given by :
\begin{align}
  &\Delta w(\Delta t)=\{
  \begin{array}{l}
A_{+}(w) e^{-\frac{|\Delta t|}{\tau_{+}}} \quad \quad \quad \text { if } \Delta t \geq 0 \\
-A_{-}(w) e^{-\frac{|\Delta t|}{\tau_{-}}}\quad \quad \quad  \text { if } \Delta t<0
\end{array}   \\
&\text{s.t.} A_{+}(w)=\eta_{+}(w_{\max }-w), A_{-}(w)=\eta_{-}(w-w_{\min }) \nonumber
\end{align}

where $\Delta t=t_{\text {post }}-t_{\text {pre }}$ is the time difference between the post-synaptic spike and the presynaptic one, with synaptic time-constant $\tau_{\pm}$.
The RSNN is made of three layers: (1) an input encoding layer ($\mathcal{I}$), (2) a recurrent spiking layer ($\mathcal{R}$), and (3) an output decoding layer ($\mathcal{O}$). The recurrent layer consists of excitatory and inhibitory neurons, distributed in a ratio of $N_{E}: N_{I}=4: 1$. The PSPs of post-synaptic neurons produced by the excitatory neurons are positive, while those produced by the inhibitory neurons are negative. We used a LIF neuron model and trained the model using STDP rules. Fig. \ref{fig:rsnn} shows an RSNN with LIF neurons and STDP synapses.

\textbf{Randomly Distributed Embedding (RDE) }\label{sec:rde} algorithm was proposed by Ma et al. \cite{ma2018randomly} to accurately predict future states based on short-term high-dimensional data. RDE accurately predicts future dynamics based on short-term high-dimensional data. To compensate for the limited number of time points, the RDE framework creates the distribution of information from the interactions among high-dimensional variables. Thus, instead of roughly predicting a  single trial of future values, RDE achieves accurate prediction using the distribution information. The RDE framework randomly generates sufficient low-dimensional non-delay embeddings from the observed data of high-dimensional variables. Each of these non-delay embeddings is then mapped to a delay embedding constructed from the data of the target variable. A target variable is a variable that we aim to predict using the RDE framework. Any of these mappings can perform as a low-dimensional weak predictor for future state prediction, and all such mappings generate a distribution of predicted future states. This distribution patches all pieces of association information from various embeddings into the complete dynamics of the target variable. After being operated by appropriate estimation strategies, it creates a stronger predictor for achieving prediction in a more reliable and robust form. Our proposed method performs the RDE-based state-space reconstruction method shown in Fig. \ref{fig:diagram} every time the batch error crosses some threshold. We use two metrics to calculate batch error: the RMSE error between the predicted and the observed time series and the Wasserstein distance between the persistent homologies of the two $d_{W,1}(\hat{p}_H, p_H)$, hereafter denoted as $d_W$. The details about the two loss metrics are further discussed in Section \ref{sec:metrics}.

\subsection{Topological Data Analysis and Persistent Homology:} Topological Data Analysis (TDA) \cite{bubenik2020persistence} uses information from topological structures in complex data for statistical analysis and learning to identify shape-like structures in data. This paper uses a mathematical tool used in TDA called persistent homology, a part of the computational topology that converts data into simplicial complexes. A simplicial complex is a space with a triangulation. Formally, a simplicial complex $K$ in $\mathbb{R}^n$ is a collection of simplices in $\mathbb{R}^n$ such that (i) every face of a simplex of $K$ is in $K$, and (ii) the intersection of any two simplices of $K$ is a face of each of them. Thus, persistent homology computes the birth and death of such topologies via a persistence diagram.
The procedure to compute persistent homology associated with the input point cloud data set involves the construction of filtration of simplicial complexes, ordered with respect to some resolution (scaling) parameter. A topological feature that persists for a more extensive range of scales is a significant one. An important reason why we use persistent homology is that it does not require an artificial cutoff between signal and noise since all topological features from the data are preserved, and weights are assigned according to their persistence. The output of this filtration procedure generates the persistence diagram. In the persistence diagram, the two coordinates of each point represent the birth value and the death value of a k-dimensional hole. A key property of persistent homology is that both persistence diagrams are robust under perturbations of the underlying data. The shift in the persistence diagram is directly proportional to any change in the dataset. These properties are leveraged for developing statistical methods for data analysis using persistent homologies. TDA provides a new type of analysis that complements standard statistical measures like RMSE. We use TDA to study the shape of the time series data by calculating $d_W$ to detect and quantify topological patterns that appear in multidimensional time series \cite{gidea2018topological}. Following the works of Gidea et al. \cite{gidea2018topological}, we apply TDA methods to explore the temporal behavior of topological features in the dynamical system time series data as the state of the dynamical system evolves. Using a sliding window, we extract time-dependent point cloud data sets to which we associate a topological space. We detect transient loops in this space and measure their persistence, encoded in real-valued functions called a persistence landscape.

\section{Methods} \label{sec:IV}

\subsection{General Framework} \label{sec:general}
There are many components in the unsupervised online learning methodology. Figure \ref{fig:diagram} visualizes the complete data flow of the model to be implemented. First, the evolving time series is derived by projecting the Lorenz63 (L63) dynamical system ~\cite{lorenz1963deterministic} onto a single axis. The L63 system works in four modes: fixed point, chaos, limit cycle, and normal. Further details about the L63 dynamical system are given in Section \ref{sec:V}a. We consider the underlying dynamical system shifting from one mode of operation to another to result in an evolving time series as depicted in Fig. \ref{fig:predictions}(a). This time series is then fed as input into a layer of encoder neurons for converting the signal into spikes using a temporal encoding scheme. These spike trains are given as input to the recurrent spiking neural network (RSNN), which is trained continually using spike timing dependent plasticity rule (STDP) \cite{chakraborty2022heterogeneous}, \cite{chakraborty2023heterogeneous}. Hence, we use the CS-HiBet method \cite{mahyar2018compressive} to efficiently detect top-k nodes in networks with the highest $\mathcal{C}_b$. CS-HiBet can perform as a distributed algorithm by using only the local information at each node, which makes it highly scalable. We record the membrane potential of these K nodes and reconstruct the underlying dynamical system from this data using random delay embedding (RDE) \cite{ma2018randomly} from which we get the final predictions.

\textbf{Encoder } encodes the time series data into a spike train. For this paper, we use a population Poisson encoder, i.e., each value of the continuous input time series signal is converted into a series of spike trains. The signal must be represented as spikes for the RSNN to process our time series to higher dimensional output. The spikes are only generated whenever the signal changes in value. This paper uses a temporal encoding mechanism to represent the signals efficiently. We implement the encoding mechanism based on the Step-Forward (SF) algorithm \cite{petro2019selection}. The property of temporal encodings to transfer required information faster than rate encoding can be helpful for online learning. Online adapting models that have to make quick decisions based on changes in the environment benefit most from temporal encoding.

\textbf{Neuron Sampling: }The proposed algorithm is based on the hypothesis that the RSNN captures the dynamics of the underlying system from the observed input time series data. Thus, a reconstruction of the dynamical system from the neuron states can predict the future states of the dynamical system. In this work, we use a neuron sampling algorithm based on $\mathcal{C}_b$ and select the neurons with the highest information flow.  
$\mathcal{C}_b$ is used to detect the degree of influence a node has over the flow of information in a network and to find nodes that serve as a bridge between two different parts of the network. Hence, we can interpret the nodes with high $\mathcal{C}_b$ similar to the bottleneck layer neurons in an autoencoder that showcase a lower-dimensional embedding of the higher-dimensional input space.
$\mathcal{C}_b$ measures the proportion of shortest paths in the network passing through a specific node. This is the fraction of times a node acts as a bridge in transferring valuable information between any pair of nodes along their shortest paths within the network. Hence, $\mathcal{C}_b$ is correlated to the nodes' importance from an information-flow standpoint in the network. Following previous works \cite{freeman1977set}, we define $\mathcal{C}_b$ of node $u \in V$ as:
$\displaystyle \mathcal{C}_b=\sum_{v, w, v \neq w} \frac{\sigma_{v w}(u)}{\sigma_{v w}}$, where $\sigma_{v w}$ is the total number of shortest paths between every $v, w \in V, v \neq w$, and $\sigma_{v w}(u)$ is the number of such paths that pass through $u$.

\textbf{Decoder}
We take the membrane potential outputs of the sampled nodes and feed them into the decoder to output a high-dimensional time series. To represent the RSNN state $r(t)$, we take the sum of all the spikes $s$ over the last $\tau$ time steps into account.
$\displaystyle
r_i^M(t)=\sum_{n=0}^\tau \gamma^n s_i(t-n) \quad \forall i \in E
$. By definition, this is an exponentially decreasing rate of decoding using a sliding window. A time series that is highly volatile in a short time frame requires the decoder to react fast to the changes. The parameters $\tau$ and $\gamma$ are balanced to optimize the memory size of the stored data and its containment of information, including adjusting $\tau$ to the pace at which the temporal data is presented processed. Since we model a continually evolving system, we chose the values of $\tau \le 50$ and $\gamma^{\tau -1} \ge 0.1$.

\subsection{Real-time Error Computation }\label{sec:metrics}  

We note that the reconstruction of the dynamical system is done every time some loss metric between the predicted and observed time series exceeds some threshold. A more frequent reconstruction would make a system inefficient and prone to noise and outliers. For the experiments, we used an RSNN with 5000 LIF neurons and a synapse connectivity probability of 20\%. However, though the model was the same for all the simulations,  we used two different loss functions for reconstructions as follows:


\textbf{RMSE-CLURSNN: }We use the RMSE loss as a metric to evaluate the loss between the predicted and the observed time series. The RMSE loss is given as $\text{RMSE}=\sqrt{\frac{1}{n} \sum_{i=1}^n\left(y_i-\hat{y}_i\right)^2}$ where $y$ is the observed time series and $\hat{y}$ is the predicted time series. We calculate the RMSE loss on a rolling window of the observed and predicted time series.

\begin{figure}
    \centering
    \includegraphics[width=\textwidth]{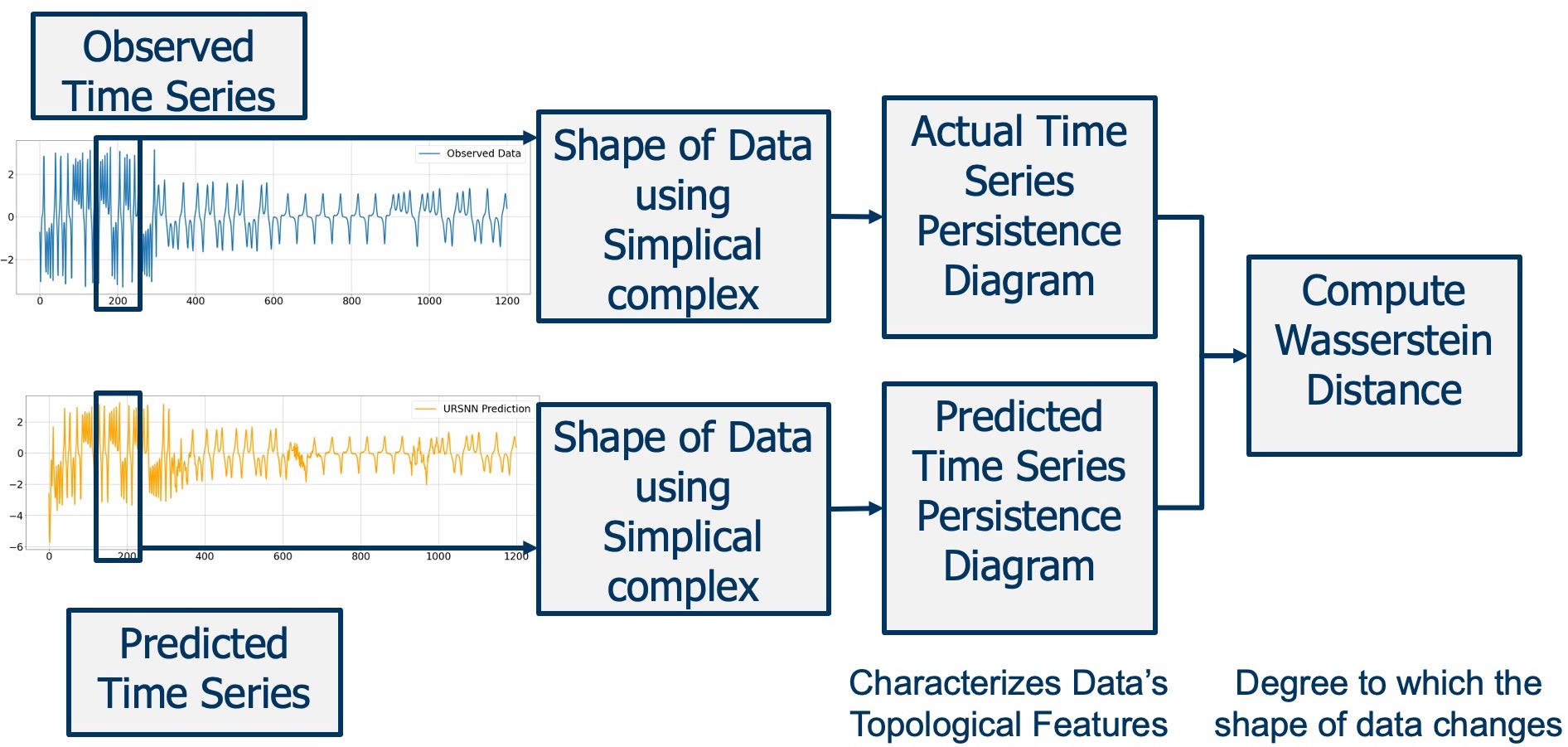}
    \caption{Block Diagram showing the computation of Wasserstein Distance between the persistent homologies of the observed and the predicted time series }
    \label{fig:block2}
\end{figure}

\textbf{Wass-CLURSNN: } Though the standard RMSE error is an efficient loss metric, it fails to capture whether the predicted time series successfully captures the \textit{shape} of the data. When using RMSE, we can get a high value even though the predicted time series captures the structure of the data. Since we aim to reconstruct the underlying dynamical system, we are more interested in whether we can capture the dynamics successfully, which might lead to an amplified or time-shifted prediction. Though these types of distortions do not change the fact that the model has successfully learned the underlying model, it leads to a high RMSE loss, which is undesirable. Hence, to bypass this problem, we introduce a novel loss metric in this paper. We introduce $d_W$ as an alternate loss metric instead of the standard RMSE error. As described before, the persistent homologies capture the topological features of the data and not just the point values, unlike RMSE. $d_W$ is calculated on a rolling time window. The detailed procedure for calculating $d_{W}$ is shown in Fig. \ref{fig:block2}. 
A persistence diagram is a multiset of points in the extended plane, $\overline{\mathbb{R}}^2$. Let $X$ and $Y$ be two persistence diagrams. To define the distance between them, we consider bijections $\eta: X \rightarrow Y$ and take the sum of $q$-th powers of the $L_{\infty}$-distances between corresponding points, and minimize overall bijections. 
Measuring the distance between points $x=\left(x_1, x_2\right)$ and $y=\left(y_1, y_2\right)$ as $\|x-y\|_{\infty}=\max \left\{\left|x_1-y_1\right|,\left|x_2-y_2\right|\right\}$ and taking the infimum over all bijections, we get the q-Wasserstein distance $d_{W.q}$ as:
$\displaystyle d_{W,q}(X, Y)=\left[\inf _{\eta: X \rightarrow Y} \sum_{x \in X}\|x-\eta(x)\|_{\infty}^q\right]^{1 / q} .$

The RSNN model is kept the same for either case; hence, the learning procedure does not change. We only use the error function as a metric for efficient reconstruction of the dynamical system attractor space based on the sampled nodal observations of the RSNN.

\section{Experiments} \label{sec:V}
\subsection{Dataset}
\textbf{ \textit{Synthetic:} } \textit{ Evolving Lorenz63 (L63) System:}  We study an evolving Lorenz system using the RSNN models. The classical three-dimensional L63 attractor ~\cite{lorenz1963deterministic} was the first example of a low-dimensional system with chaotic solutions. The following differential equations describe the L63 system:
\begin{equation}
   \frac{\mathrm{d} x}{\mathrm{~d} t}=\sigma(y-x), \frac{\mathrm{d} y}{\mathrm{~d} t}=x(\rho-z)-y, \frac{\mathrm{d} z}{\mathrm{~d} t}=x y-\beta z 
\end{equation}

We consider four different types of behavior of the L63 system depending on the parameter values:
\begin{itemize}
    \item \textit{Fixed Points } (F.P) $(\rho=60, \sigma=20, \beta=8)$
    \item \textit{Chaos } $(\rho=36,  \sigma=8.5, \beta=3.5)$
    \item \textit{Limit Cycles } (L.C.) $(\rho=35, \sigma=21, \beta=1)$
    \item \textit{Normal } $(\rho=28, \sigma=10, \beta=2.66)$
\end{itemize}

\textbf{\textit{Real-world Datasets: }} \textbf{YReal Datasets: } \cite{laptev2015s5}  The YReal time series is the real-world timeseries from the A1Benchmark data in Yahoo's S5 dataset. We use the time series in A4Benchmark containing synthetic outliers and change points. The time series represents the metrics of various Yahoo services with different change points and outliers. Since our model is learning from streaming data, the model's performance on outliers can help us determine the robustness of the methods. 
\par \textbf{Dow Jones Industrial Average: }We use the daily index of the Dow Jones Industrial Average (DJIA) from 1885–1962.

\subsection{Methodology}

\begin{figure*}
        \centering
    \includegraphics[width=\textwidth]{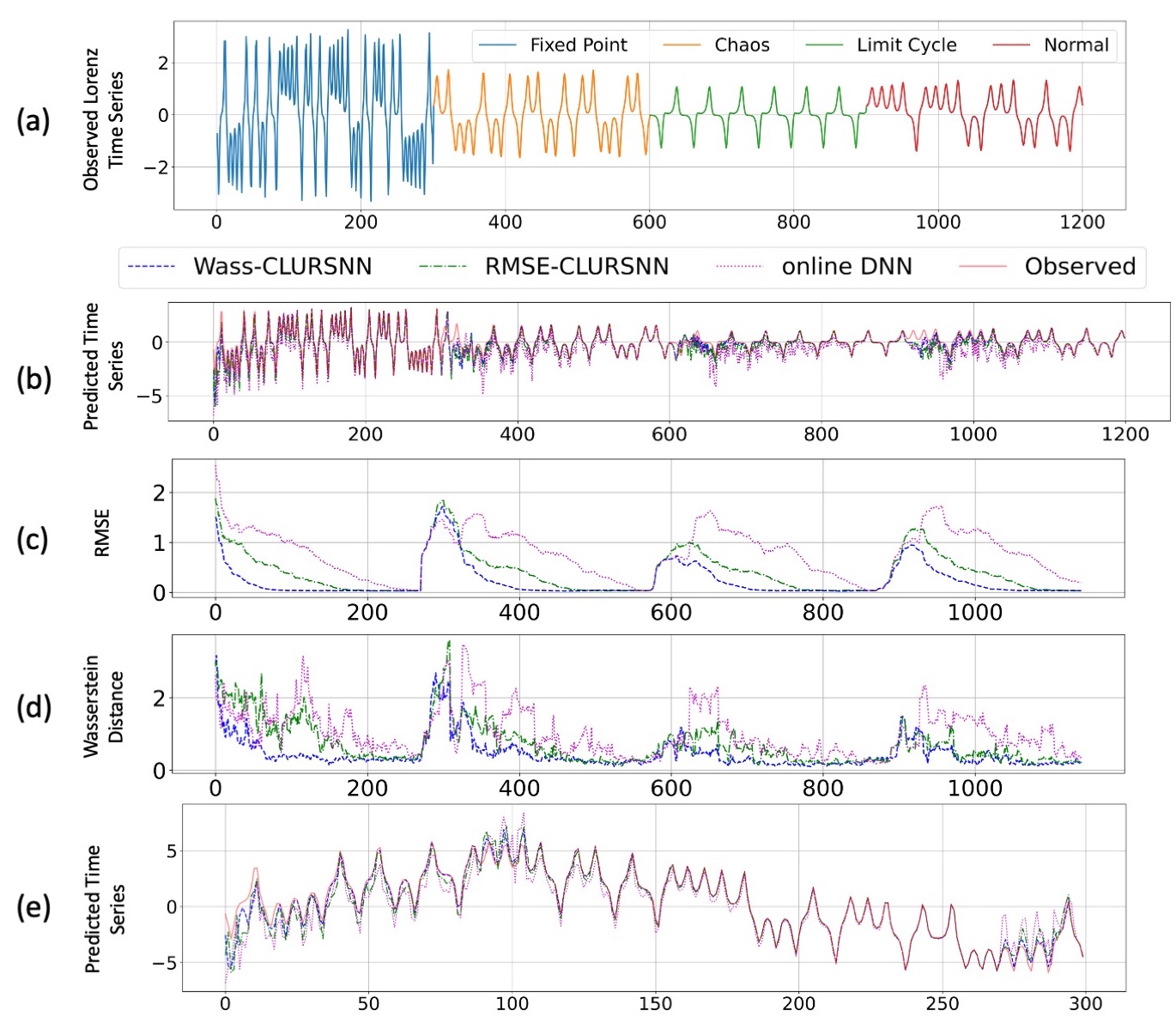}
    \caption{(a) Fig showing the four modes of operation of the L63 dynamical system. Each mode has 300 timesteps. (b) The predicted and the observed time series data for the three models compared in the paper - RMSE-CLURSNN, Wass-CLURSNN, and online DNN. (c) The RMSE loss between the predicted and the observed time series computed in a rolling time window of 30 timesteps for the three models (d) $d_{W}$ of the three models (e) The predicted and observed time series of the time series with a sinusoidal trend. We only show the first 300 timesteps (corresponding to the FP of the L63 System). A similar process was repeated for the other three modes of operation. }
    \label{fig:predictions}
\end{figure*}

\textbf{Baselines DNN Model:} We compare the performance of the novel SNN-based unsupervised learning methodology with a supervised DNN-based online learning model. We choose the conventional real-time current learning (RTRL) of LSTM neural networks\cite{williams1989learning}, henceforth called online DNN. It updates the network using the newly available data without considering the change points. The online DNN model uses a stochastic gradient descent (SGD) based method to learn the model from the
streaming time series. With new observations, the model
parameters are updated online according to the
gradients of the loss of the newly available data. Online DNN explores the local features of time series
to weigh the learning method's gradients with the local data's distributional properties. In particular, if a newly
available observation is a potential outlier behaving differently
from the regular patterns, the corresponding gradient will be
down-weighted to avoid abruptly leading the online model to drift from the underlying patterns.

\subsection{Results}

\textbf{Prediction Performance: } We compare the online prediction performances of the RSNN models with the baseline DNN models. Fig. \ref{fig:predictions}(b) shows the prediction results of the Wass-CLURSNN, RMSE-CLURSNN, and the online DNN models compared to the observed time series. The Wass-CLURNN can follow the evolving system better than the other two models. The online DNN model shows the worst convergence among the three. These results are further validated by the plot of RMSE and $d_{W}$ shown in Fig. \ref{fig:predictions}(c, d), respectively. From these plots, we see that the Wass-CLURSNN performs the best and converges the fastest among the three, while the online DNN, i.e., the LSTM model, performs the worst and converges the slowest. It is to be noted here that the RMSE and $d_{W}$ are calculated here on a rolling window of 30 timesteps.

\textbf{Adaptation to Timeseries with Sinusoidal Trend: } 
We evaluate the rate at which the different methods used in this paper can adapt to a changing dynamical system. To make the system more complex and study this adaptation behavior further, we use a sinusoidal signal as a trend line for the time series from the L63 system sinusoidal oscillating within a sinusoidal function. We call this process the \textit{sinusoidal trend of the time series}. The frequency of the trendline sinusoidal signal determines the time series' evolution rate, where a higher frequency (smaller Time period) implies the time series changes rapidly and the model needs to adapt more quickly to the changing system.
Studying the prediction performance of the three models (Wass-CLURSNN, RMSE-CLURSNN, and online DNN) on this sine-modulated L63 time series, we test how each of these models can adapt to changing dynamical systems. Fig \ref{fig:predictions}(e) shows an instance of the first 300 timesteps of the sine-modulated time series. The figure shows where the L63 system in the fixed point mode is modulated using a sine wave of the time period of 300 timesteps and an amplitude of 3. Table \ref{tab:results} shows the average RMSE and average $d_{W}$ for each of the three methods. The mean and standard deviation are taken over each of the four modes of operation. First, we report the baseline case for an unmodulated time series. We also give the results of sine-waves with periods of 100, 300, and 500 with amplitude of 3 and 5 for each case. The results are summarized in Table \ref{tab:results}. We see that Wass-CLURSNN outperforms RMSE-CLURSNN and online DNN models.

\begin{table*}[]
\centering
\caption{Table comparing the performances of the three models - RMSE-CLURSNN, Wass-CLURSNN, and online DNN when using different sine-modulated time series of the evolving L63 dynamical system. The experiment is repeated for each of the four modes of operation, and the mean and standard deviation are calculated from these observations.}
\label{tab:results}
\resizebox{\textwidth}{!}{%
\begin{tabular}{|c|c|cc|cc|cc|}
\hline
\multirow{2}{*}{\textbf{\begin{tabular}[c]{@{}c@{}}Trendline\\ Sine Amplitude\\ (A)\end{tabular}}} &
  \multirow{2}{*}{\textbf{\begin{tabular}[c]{@{}c@{}}Trendline\\ Sine \\ Time Period\end{tabular}}} &
  \multicolumn{2}{c|}{\textbf{RMSE-CLURSNN}} &
  \multicolumn{2}{c|}{\textbf{Wass-CLURSNN}} &
  \multicolumn{2}{c|}{\textbf{online DNN}} \\ \cline{3-8} 
 &
   &
  \multicolumn{1}{c|}{\textbf{\begin{tabular}[c]{@{}c@{}}Average\\  RMSE\end{tabular}}} &
  \textbf{\begin{tabular}[c]{@{}c@{}}Average \\ $d_{W}$ \end{tabular}} &
  \multicolumn{1}{c|}{\textbf{\begin{tabular}[c]{@{}c@{}}Average\\  RMSE\end{tabular}}} &
  \textbf{\begin{tabular}[c]{@{}c@{}}Average \\ $d_{W}$\end{tabular}} &
  \multicolumn{1}{c|}{\textbf{\begin{tabular}[c]{@{}c@{}}Average\\  RMSE\end{tabular}}} &
  \textbf{\begin{tabular}[c]{@{}c@{}}Average \\ $d_{W}$\end{tabular}} \\ \hline
\textbf{0} &
  \textbf{-} &
  \multicolumn{1}{c|}{$0.324 \pm 0.021$} &
  \textit{$0.301 \pm 0.015$} &
  \multicolumn{1}{c|}{$0.206 \pm 0.02$} &
  \textit{$0.166 \pm 0.013$} &
  \multicolumn{1}{c|}{$0.913 \pm 0.064$} &
  \textit{$0.628 \pm 0.044$} \\ \hline
\multirow{3}{*}{\textbf{3}} &
  \textbf{100} &
  \multicolumn{1}{c|}{$0.615 \pm 0.037$} &
  \textit{$0.352 \pm 0.019$} &
  \multicolumn{1}{c|}{$0.441 \pm 0.031$} &
  \textit{$0.198 \pm 0.013$} &
  \multicolumn{1}{c|}{$1.554 \pm 0.11$} &
  \textit{$0.954 \pm 0.067$} \\ \cline{2-8} 
 &
  \textbf{300} &
  \multicolumn{1}{c|}{$0.596 \pm 0.031$} &
  \textit{$0.326 \pm 0.014$} &
  \multicolumn{1}{c|}{$0.322 \pm 0.024$} &
  \textit{$0.183 \pm 0.011$} &
  \multicolumn{1}{c|}{$1.388 \pm 0.094$} &
  \textit{$0.799 \pm 0.051$} \\ \cline{2-8} 
 &
  \textbf{500} &
  \multicolumn{1}{c|}{$0.357 \pm 0.025$} &
  \textit{$0.311 \pm 0.011$} &
  \multicolumn{1}{c|}{$0.217 \pm 0.019$} &
  \textit{$0.174 \pm 0.01$} &
  \multicolumn{1}{c|}{$1.068 \pm 0.075$} &
  \textit{$0.701 \pm 0.042$} \\ \hline
\multirow{3}{*}{\textbf{5}} &
  \textbf{100} &
  \multicolumn{1}{c|}{$0.719 \pm  0.053$} &
  \textit{$0.434 \pm 0.021$} &
  \multicolumn{1}{c|}{$0.507 \pm 0.048$} &
  \textit{$0.276 \pm 0.019$} &
  \multicolumn{1}{c|}{$1.994 \pm 0.139$} &
  \textit{$0.973 \pm 0.061$} \\ \cline{2-8} 
 &
  \textbf{300} &
  \multicolumn{1}{c|}{$0.655 \pm 0.049$} &
  \textit{$0.388 \pm 0.019$} &
  \multicolumn{1}{c|}{$0.389 \pm 0.044$} &
  \textit{$0.221 \pm 0.016$} &
  \multicolumn{1}{c|}{$1.635 \pm 0.111$} &
  \textit{$0.882 \pm 0.598$} \\ \cline{2-8} 
 &
  \textbf{500} &
  \multicolumn{1}{c|}{$0.433 \pm 0.032$} &
  \textit{$0.335 \pm 0.016$} &
  \multicolumn{1}{c|}{$0.294 \pm 0.029$} &
  \textit{$0.195 \pm 0.014$} &
  \multicolumn{1}{c|}{$1.121 \pm 0.079$} &
  \textit{$0.866 \pm 0.554$} \\ \hline
\end{tabular}%
}
\end{table*}

\begin{figure}
    \centering
    \includegraphics[width=\textwidth]{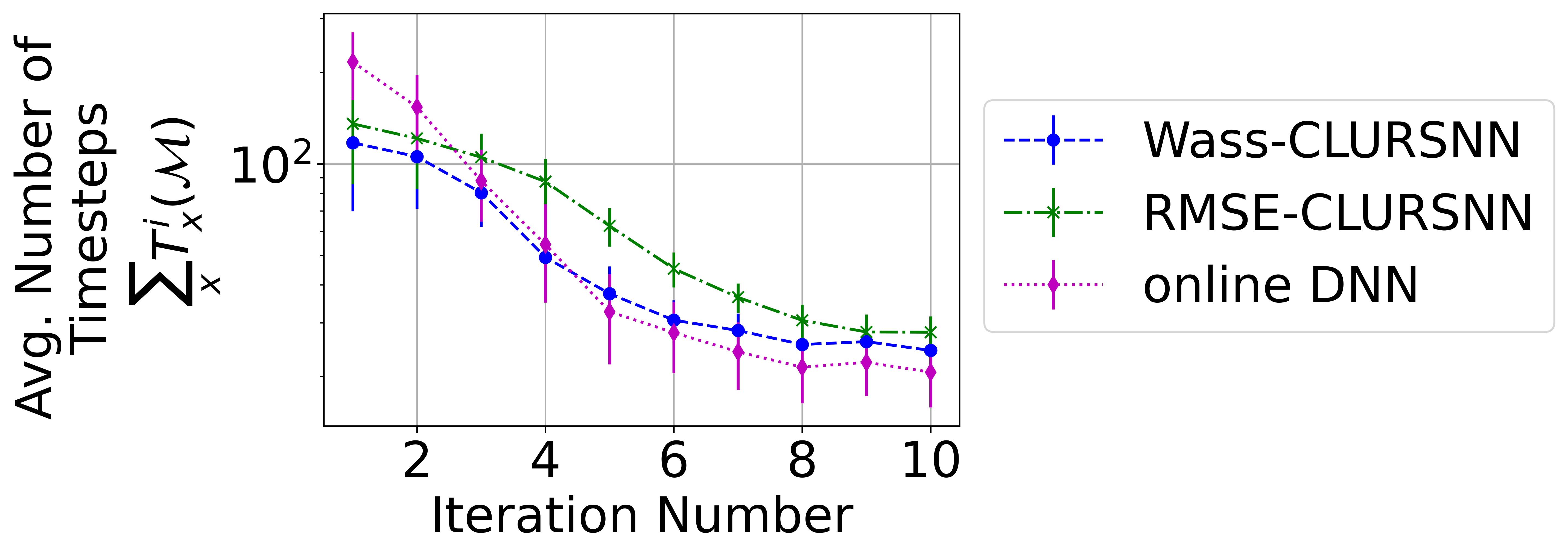}
    \caption{Figure compares the convergence ability of the three models over repeated iterations of the same input. The convergence is calculated using the mean number of timesteps required for the model to converge in each case. The experiment was repeated for each of the four modes of operation of the L63 system, and the mean and standard deviation from these observations are plotted. We use the log-scale value for the y-axis. }
    \label{fig:conv}
\end{figure}

\textbf{Memory Retention and Convergence: } We know that the time series constitutes four different modes of operation of the L63 dynamical system. The time series observed for these different modes of operation is shown in Fig. \ref{fig:predictions}(a). Let us denote these behaviors as $a^x$ where $x$ can be $FP = $Fixed Point , $C$= Chaotic, $LC$=Limit cycles, $N$= Normal. Also, let us denote the prediction time series as $a_n$. We say that $a_n$ has converged to a value L if $\forall \varepsilon >0 \exists N $ such that $n \ge N \Rightarrow |a-L| < \varepsilon$. For our simulations, we fix the value of $\varepsilon = 0.1$. We use a single training epoch for the supervised models to keep the results comparable to unsupervised SNN models. Hence, to see how the convergence behavior changes when the same input is repeated, we perform the following experiment: 
since each iteration constitutes four modes of operation, we take 300 timesteps of each mode and calculate the number of timesteps the models require to converge for each mode. Let the number of timesteps required for iteration $i$ for mode $x$ by the model $\mathcal{M}$ be denoted by $T_{x}^{i}(\mathcal{M})$. 
Hence, we compare the average of the number of timesteps required to converge over all the modes of operation for each iteration, i.e., $ \sum_{x} T_{x}^{i}(\mathcal{M})$. The results are plotted in Fig. \ref{fig:conv}. The figure shows that the Wass-CLURSNN outperforms the other methods in the first few iterations and converges the fastest. We also see that the online DNN model performs the worst among the three, indicating that when training, the DNN-based model requires more training samples to converge compared to the unsupervised RSNN-based models. However, as the number of iterations increases, the online DNN takes much fewer timesteps. From this observation, we can infer that DNN-based models can converge much faster once the models are fully trained. Thus, the DNN-based models fail when the underlying dynamical system is fast evolving and there is not a lot of data available, and that is when the RSNN-based models perform the best. Thus, these two models have two separate regions of operations.

\begin{table}[]
\centering
\resizebox{\textwidth}{!}{%
\begin{tabular}{|c|cc|cc|}
\hline
\multirow{2}{*}{\backslashbox{\textbf{Metric}}{\textbf{Model}}} & \multicolumn{2}{c|}{\textbf{RMSE-CLURSNN}} & \multicolumn{2}{c|}{\textbf{Wass-CLURSNN}} \\ \cline{2-5} 
 & \multicolumn{1}{c|}{\begin{tabular}[c]{@{}c@{}}\textbf{\textit{All}} \\ \textbf{\textit{Neurons}}\end{tabular}} & \begin{tabular}[c]{@{}c@{}}\textbf{\textit{Sampled}} \\ \textbf{\textit{Neurons}}\end{tabular} & \multicolumn{1}{c|}{\begin{tabular}[c]{@{}c@{}}\textbf{\textit{All}} \\ \textbf{\textit{Neurons}}\end{tabular}} & \begin{tabular}[c]{@{}c@{}}\textbf{\textit{Sampled}} \\ \textbf{\textit{Neurons}}\end{tabular} \\ \hline
\begin{tabular}[c]{@{}c@{}}\textbf{\textit{RMSE}}\\ (Mean $\pm$ Std. Deviation)\end{tabular} & \multicolumn{1}{c|}{\begin{tabular}[c]{@{}c@{}}$0.544$ \\ $\pm 0.112$\end{tabular}} & \begin{tabular}[c]{@{}c@{}}$0.324$ \\ $\pm 0.021$\end{tabular} & \multicolumn{1}{c|}{\begin{tabular}[c]{@{}c@{}}$0.411$\\ $\pm 0.132$\end{tabular}} & \begin{tabular}[c]{@{}c@{}}$0.206$\\ $\pm 0.02$\end{tabular} \\ \hline
\begin{tabular}[c]{@{}c@{}}\textbf{\textit{Wasserstein Distance}}\\ (Mean $\pm$ Std. Deviation)\end{tabular} & \multicolumn{1}{c|}{\begin{tabular}[c]{@{}c@{}}$0.457$ \\ $\pm 0.854$\end{tabular}} & \begin{tabular}[c]{@{}c@{}}$0.301$\\ $\pm 0.015$\end{tabular} & \multicolumn{1}{c|}{\begin{tabular}[c]{@{}c@{}}$0.351$ \\ $\pm 0.773$\end{tabular}} & \begin{tabular}[c]{@{}c@{}}$0.166$\\ $\pm 0.013 $\end{tabular} \\ \hline
\begin{tabular}[c]{@{}c@{}}\textbf{\textit{Timesteps to}} \\ \textbf{\textit{Convergence}}\\ (Mean $\pm$ Std. Deviation)\end{tabular} & \multicolumn{1}{c|}{\begin{tabular}[c]{@{}c@{}}$178.39$\\ $\pm 46.75$\end{tabular}} & \begin{tabular}[c]{@{}c@{}}$135.45$ \\ $\pm 29.68$\end{tabular} & \multicolumn{1}{c|}{\begin{tabular}[c]{@{}c@{}}$153.15$ \\ $\pm 41.34$\end{tabular}} & \begin{tabular}[c]{@{}c@{}}$117.38$\\ $\pm 27.32$\end{tabular} \\ \hline
\end{tabular}%
}
\caption{Table comparing the RMSE loss, $d_{W}$ and Timesteps required to converge for (i)RMSE-CLURSNN and (ii)Wass-CLURSNN using all the neurons in the recurrent layer vs. sampling neurons with highest $\mathcal{C}_b$. The experiment was repeated for the 4 modes of operation of L63 system, and the mean and standard deviation is calculated from these observations.}
\label{tab:sampling}
\end{table}

\textbf{Node Sampling: } We evaluate the model's performance using all the neurons in the recurrent layer compared to the case when we are sampling neurons based on $\mathcal{C}_b$. As discussed earlier, the nodes are sampled according to the ones with the highest $\mathcal{C}_b$ \cite{mahyar2018compressive}. We calculate these two types of predictions' average RMSE and average $d_{W}$. We calculate the mean and standard deviations over the four modes of operation of the L63 system. The results are summarized in Table \ref{tab:sampling}. We observe that both the models perform better and converge faster, with lesser standard deviation when using the sampled neurons than all the neurons in the recurrent layer. We may interpret this result as follows: since $\mathcal{C}_b$ captures a node's role in allowing information to pass from one part of the network to the other, it behaves like the bottleneck layer in an autoencoder. Hence sampling neurons with the highest $\mathcal{C}_b$, we can better capture the dynamical system's latent dynamics, leading to better performance.

\textbf{Performance on Real-world Datasets: } We compared the performances of the different models on some real-world datasets like YReal, and DJIA (as described above). The results are tabulated in Table \ref{tab:real}. The proposed methods RMSE-CLURSNN and Wass-CLURSNN outperform the other supervised DNN models.

\begin{table}[]
\centering
\caption{Table comparing the RMSE performance of different models on real-world datasets}
\label{tab:real}
\resizebox{\textwidth}{!}{%
\begin{tabular}{|cc|c|c|}
\hline
\multicolumn{2}{|c|}{\backslashbox{\textbf{Models}}{\textbf{Dataset}}}                                                  & \textbf{YReal} & \textbf{DJIA} \\ \hline
\multicolumn{1}{|c|}{\multirow{1}{*}{\textit{\begin{tabular}[c]{@{}c@{}}DNN\\ (Supervised)\end{tabular}}}} &
  \textit{\textbf{SR-LSTM}} &
  
  13816.99 &
  1.76 \\ \cline{2-4} 
\multicolumn{1}{|c|}{} & \textit{\textbf{\begin{tabular}[c]{@{}c@{}}RLSTM\\ (online DNN)\end{tabular}}}         & 10104.24       & 1.85        \\ \hline
\multicolumn{1}{|c|}{\multirow{1}{*}{\textit{\begin{tabular}[c]{@{}c@{}}SNN\\ (Unsupervised)\end{tabular}}}} &
  \textit{\textbf{\begin{tabular}[c]{@{}c@{}}RMSE-CLURSNN\\ (Ours)\end{tabular}}} &
  8938.25 &
  1.41 \\ \cline{2-4} 
\multicolumn{1}{|c|}{} & \textit{\textbf{\begin{tabular}[c]{@{}c@{}}Wass-CLURSN\\ (Ours)\end{tabular}}}         & 8875.73        & 1.31        \\ \hline
\end{tabular}%
}
\end{table}

\section{Conclusions} \label{sec:VI}

This paper proposes a novel methodology for online prediction of the time series from an evolving dynamical system using a completely unsupervised bio-plausible recurrent spiking neural network model trained with spike-timing-dependent plasticity. We leveraged recent works on topological data analysis and embedding theory to develop a new methodology for continuous reconstruction of the underlying dynamical system based on membrane potential measurements of nodes in the RSNN with the highest betweenness centrality, which behaves similarly to the bottleneck layer in an autoencoder. We proposed the RMSE-CLURSNN and the Wass-CLRSNN model, where the prediction error is calculated using the RMSE error and the Wasserstein distance between the persistent homologies of the predicted and observed time series $d_{W}$, respectively. We compared this new methodology's performance and convergence behavior with the online DNN model. We observed that the proposed method outperforms the standard DNN model in terms of prediction performance and convergence for an evolving dynamical system with few data points for training. However, with repeated iterations, we see that DNN-based models converge faster and show better performance. Thus, these DNN models need a large amount of data to learn. RSNN-based models, on the other hand, can adapt very quickly to changing environments.

\section*{Acknowledgement}
This work is supported by the Army Research Office and was accomplished under Grant Number W911NF-19-1-0447. The views and conclusions contained in this document are those of the authors and should not be interpreted as representing the official policies, either expressed or implied, of the Army Research Office or the U.S. Government.

\bibliographystyle{unsrt}
\bibliography{references}
\end{document}